\def\year{2020}\relax
\documentclass[letterpaper]{article} 
\usepackage{aaai20}  
\usepackage{times}  
\usepackage{helvet} 
\usepackage{courier}  
\usepackage[hyphens]{url}  
\usepackage{graphicx} 
\usepackage{graphicx}

\usepackage{array}
\newcolumntype{L}[1]{>{\raggedright\let\newline\\\arraybackslash\hspace{0pt}}m{#1}}
\newcolumntype{C}[1]{>{\centering\let\newline\\\arraybackslash\hspace{0pt}}m{#1}}
\newcolumntype{R}[1]{>{\raggedleft\let\newline\\\arraybackslash\hspace{0pt}}m{#1}}

\nocopyright

\usepackage{todonotes} 
\usepackage{color, colortbl} 
\definecolor{lightgray}{gray}{0.9}
\usepackage[utf8]{inputenc}
\usepackage[T2A,T1]{fontenc} 
\usepackage[russian,english]{babel} 
\usepackage{multirow}
\newcommand{\dragonfly}{Dragonfly\ }

\title{Improving Neural Named Entity Recognition with Gazetteers}
\author  {Chan Hee Song  \\ University of Notre Dame \\ csong1@nd.edu
\And Dawn Lawrie \\ HLTCOE, JHU\\ lawrie@jhu.edu
\And Tim Finin \\ UMBC, HLTCOE\\ finin@umbc.edu
\And James Mayfield \\ HLTCOE, JHU\\ mayfield@jhu.edu
}

\begin{document}

\maketitle

\begin{abstract}
The goal of this work is to improve the performance of
a neural named entity recognition system by adding input features that indicate a word is part of a name included in a gazetteer.
This article describes how to generate gazetteers from the Wikidata knowledge graph as well as how to integrate the
information into a neural NER system.
Experiments reveal that the approach yields performance gains in two distinct languages:
a high-resource, word-based language, English and a high-resource, character-based language, Chinese.
Experiments were also performed in a low-resource language, Russian on
a newly annotated Russian NER corpus from Reddit tagged with four core types and twelve extended types. This article reports  a baseline score. It is a longer version of a paper in the 33rd FLAIRS conference \cite{ChanHeeSong2020flairs}.
\end{abstract}

\section{Introduction}

Named Entity Recognition (NER) is an important task in natural language understanding that entails spotting mentions of conceptual entities in text and classifying them according to a given set of categories. 
It is particularly useful for downstream tasks such as  information retrieval, question answering, and knowledge graph population. 
Developing a well-performing, robust NER system can facilitate more sophisticated queries that involve entity types in information retrieval and more complete extraction of information for knowledge graph population. 

Various approaches exist to automated named entity recognition. 
Older statistical methods use conditional random fields~\cite{finkel2005}, perceptrons~\cite{ratinov2009}, and support vector machines~\cite{svmlattice}. 
More recent approaches have applied deep neural models, beginning with \citeauthor{Collobert2011}~(\citeyear{Collobert2011}).
Further advances came from the addition of a BiLSTM model with CRF 
decoding~\cite{huang2015bidirectional}, which led to the current state-of-the-art model.

Our NER architecture combines recent advances in transfer learning~\cite{devlin2018bert} and a BiLSTM-CRF model, producing a BERT-BiLSTM-CRF model.
In our model, BERT generates an embedding for each word.
This embedding is fed into a multi-layer BiLSTM, which is often jointly trained with a pre-trained encoder at training time. This fine-tunes the encoder to the NER task.
At test time, the BiLSTM outputs are decoded using a CRF.
Other approaches show similar results, such as a BiLSTM-CRF that uses a character-level CNN added to BiLSTM-CRF~\cite{Ma2016,chiu2016}. 
We adopt a BERT-based model as the baseline system for comparison.

In the context of natural language understanding, a gazetteer is simply a collection of common entity names typically organized by their entity type.  These have been widely used in natural language processing systems since the early 1990s, such as a large list of English place names provided by the MUC-5 Message Understanding Conference \cite{sundheim1993tipster} to support its TIPSTER information extraction task.  Initially these lists were employed to help recognize and process mentions of entities that were places or geo-political regions, hence the name gazetteer.  Their use quickly evolved to cover more entity types and subtypes, such as cities, people, organizations, political parties and religions.
 
Statistical approaches have benefited by using
gazetteers as an additional source of information, often
because the amount of labeled data for training an NER system tends to be small.
A lack of training data is of particular concern when using neural architectures, which generally require large amounts of training
data to perform well.
Gazetteers are much easier to produce than labeled training data and can be mined from existing sources. 
Therefore, it is important to know whether this rich source of information can
be effectively integrated into a neural model.

This paper first focuses on generating gazetteers from Wikidata,
presenting a simple way to gather a large quantity of annotated entities from Wikidata. It then
describes how to integrate the gazetteers with a neural architecture
by generating features from gazetteers alongside the features from BERT as input to the BiLSTM.
We aim to provide an additional external knowledge base to neural systems similar to the way people use external knowledge to determine what is an entity and to what category it belongs.

Adding gazetteer features (often called lexical features) to neural systems has been shown to improve performance on well-studied datasets like English OntoNotes and CONLL-2003 NER using a closed-world neural system (\emph{i.e.,} BiLSTM-CRF)~\cite{chiu2016}.
We extended this approach and validated that gazetteer features are still beneficial to datasets in a more diverse set of languages and with models that use a pre-trained encoder. 

For generality, we applied and evaluated our approaches on datasets in three languages: a high-resource, word-based language, English;
a high-resource, character-based language, Chinese;
and a lower-resource, high morphological  language, Russian.
We will first present how our gazetteer is generated from publicly available data source, Wikidata.
Then we will analyze our experimental results.

\begin{table}[t]
\centering \scriptsize
\setlength{\tabcolsep}{4pt} 
\renewcommand{\arraystretch}{1.25} 
\begin{tabular}{|l|L{2.7cm}|L{4.0cm}|} \hline
\rowcolor{lightgray} \textbf{\normalsize Type}  & \textbf{\normalsize Description} & \textbf{\normalsize Examples}\\ \hline
\textbf{PER} & Person & Enrico Rastelli, Beyoncé\\ \hline
\textbf{ORG} & Organization & International Jugglers Association\\ \hline
COMM & Commercial Org. & Penguin Magic, Papermoon Diner\\ \hline
POL & Political Organization & Green Party, United Auto Workers\\ \hline
\textbf{GPE} & Geo-political Entity & Morocco, Carlisle, Côte d'Ivoire\\ \hline
\textbf{LOC} & Natural Location & Kartchner Caverns, Lake Erie\\ \hline
FAC & Facility & Stieff Silver Building, Hoover Dam\\ \hline
GOVT & Government Building & White House, 10 Downing St.\\ \hline
AIR & Airport & Ninoy Aquino International, BWI\\ \hline
EVNT & Named Event & WWII, Operation Desert Storm\\ \hline
VEH & Vehicle & HMS Titanic, Boeing 747\\ \hline
COMP & Computer Hard/Software & Nvidia GeForce RTX 2080 Ti, Reunion\\ \hline
MIL & Military Equip. & AK-47, Fat Man, cudgel\\ \hline
MIL\_G  & Generic Military Equip. & tank, aircraft carrier, rifle \\ \hline
MIL\_N  & Named Military Equip.& USS Nimitz, 13"/35 caliber gun \\ \hline
CHEM & Chemical & Iron, NaCl, hydrochloric acid\\ \hline
MISC & Other named entity & Dark Star, Lord of the Rings\\ \hline
\end{tabular}
\caption{We worked with types where training data was available in several languages, including four core types (in bold) and twelve additional ones.} 
\label{tab:types}
\vspace{-1em}
\end{table}
\renewcommand{\arraystretch}{1} 

\section{Related Work}

Our work builds on the neural approach to NER, which
was introduced when \citeauthor{Hammerton2003}~(\citeyear{Hammerton2003}) used Long Short-Term Memory (LSTM), achieving just above average performance for English and improvement for German.  LSTM was  proposed by \citeauthor{Hochreiter:1997:LSM:1246443.1246450}~(\citeyear{Hochreiter:1997:LSM:1246443.1246450}), expanded by  \citeauthor{Gers2000}~(\citeyear{Gers2000}), and reached its modern form with \citeauthor{graves2005framewise}~(\citeyear{graves2005framewise}). Recent NER systems have adopted a forward-backward LSTM or BiLSTM, mainly using the BiLSTM-CRF architecture first proposed by \citeauthor{huang2015bidirectional}~(\citeyear{huang2015bidirectional}), and now widely studied and augmented. 
For example, \citeauthor{chiu2016}~(\citeyear{chiu2016}) and \citeauthor{ma-hovy}~(\citeyear{ma-hovy}) augmented the BiLSTM-CRF architecture with a character-level CNN 
to add  additional features to the architecture.

Adding lexical features to the system has been studied widely,
mainly by matching words in the dataset to words in pre-gathered gazetteers. \citeauthor{passos-etal-2014-lexicon}~(\citeyear{passos-etal-2014-lexicon}) uses gazetteers during embedding generation; \citeauthor{chiu2016}~(\citeyear{chiu2016}) uses gazetteers to generate a one-hot encoded match of the words in the data to those in the gazetteers; and \citeauthor{Ghaddar2016}~(\citeyear{Ghaddar2016}) generates gazetteer embeddings from Wikipedia. \citeauthor{ding-etal-2019-neural}~(\citeyear{ding-etal-2019-neural}) presents an architecture incorporating gazetteer information for Chinese, which is a language that often has a greater number of false positive matches because it is logographic. 

However, our approach provides a simple augmentation to existing neural models and demonstrates that Chinese can benefit from gazetteer matches. We take the \citeauthor{chiu2016}~(\citeyear{chiu2016})'s approach to matching the gazetteer because of its simplicity and universality in application to many different neural models. We also show that it is applicable to neural models with a deep pre-trained encoder.

Transfer learning architectures have shown significant improvement in various natural language processing tasks such as understanding, inference, question answering, and machine translation. BERT~\cite{devlin2018bert}, uses stacked bi-directional transformer layers trained on masked word prediction and next sentence prediction tasks. BERT is trained on over 3.3 billion words gathered mainly from Wikipedia and Google Books. By adding a final output layer, BERT can be adapted to many different natural language processing tasks. In this work, we apply BERT to NER and use BiLSTM-CRF as the output layer of BERT. Our approach embodies a simple architecture that does not require a dataset-specific architecture or feature engineering. 

\section{Gazetteer Creation} 

We describe the knowledge source we use to create our gazetteers and outline the process we used to automatically produce cleaned gazetteers for the entity types of interest.
\begin{figure}
\centering
{\setlength{\fboxsep}{0pt}
\setlength{\fboxrule}{1pt}
\fbox{\includegraphics[width=\columnwidth]{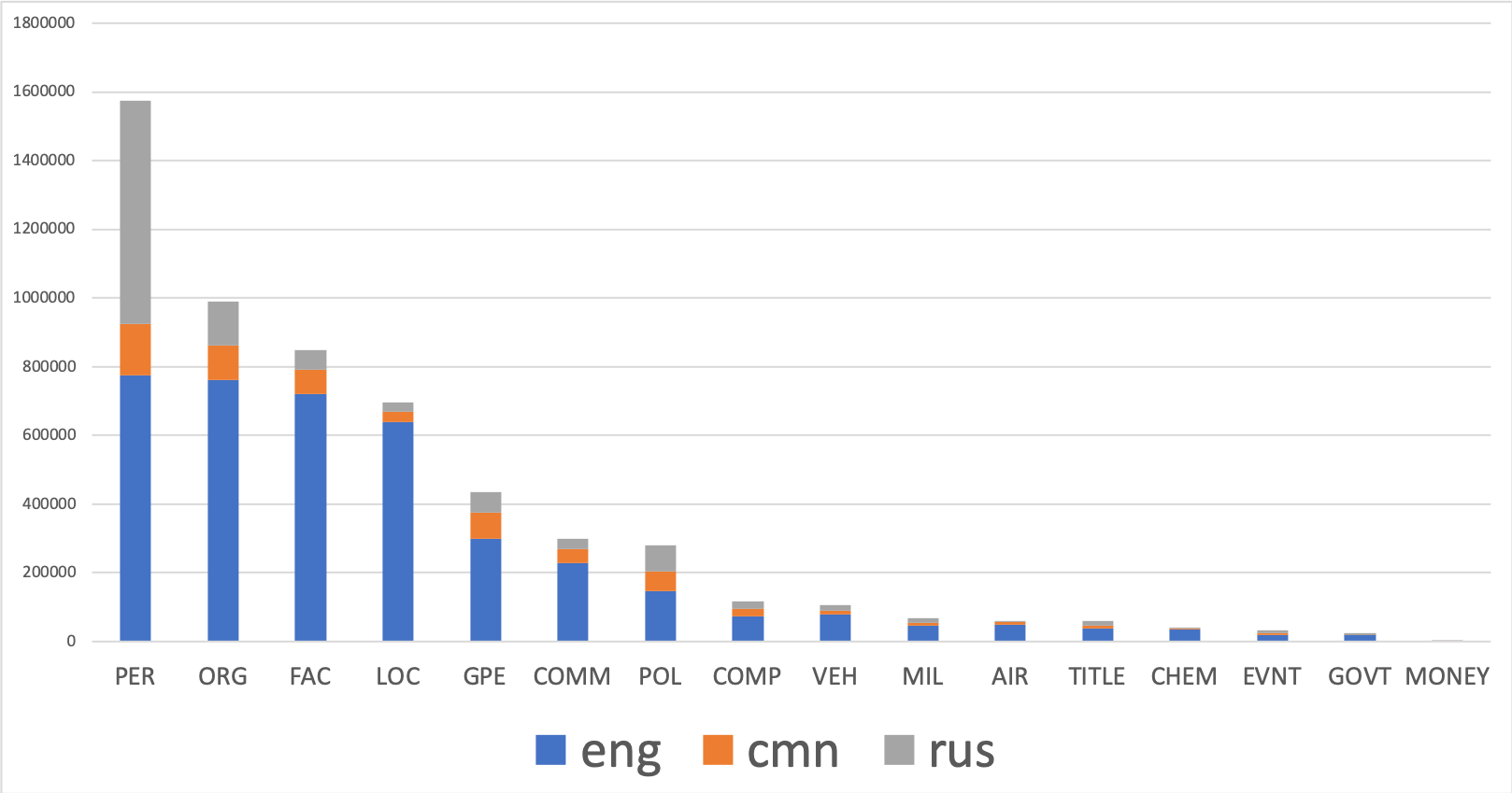}}}
\caption{Statistics for canonical names for Wikidata entities for each type.  Additional lists hold aliases and, for Russian, inflected forms. }
\end{figure}

\subsection{Wikidata Knowledge Graph}

Our gazetteers were created by extracting canonical names (e.g., Manchester United F.C.) and aliases (e.g., Red Devil, Man U) of entities of a given type (e.g., ORG) from Wikidata \cite{vrandecic:2014}.
Wikidata is a large, collaboratively edited knowledge graph with information drawn from and used by a number of Wikimedia projects,
including 310 Wikipedia sites in different languages.
Its goal is to integrate entities and knowable facts about them for use in Wikimedia sites in a language-independent manner.  Wikidata is multilingual, with all of its strings tagged with a two-letter ISO 639-1 language code.

\begin{figure}  
\includegraphics[width=\columnwidth]{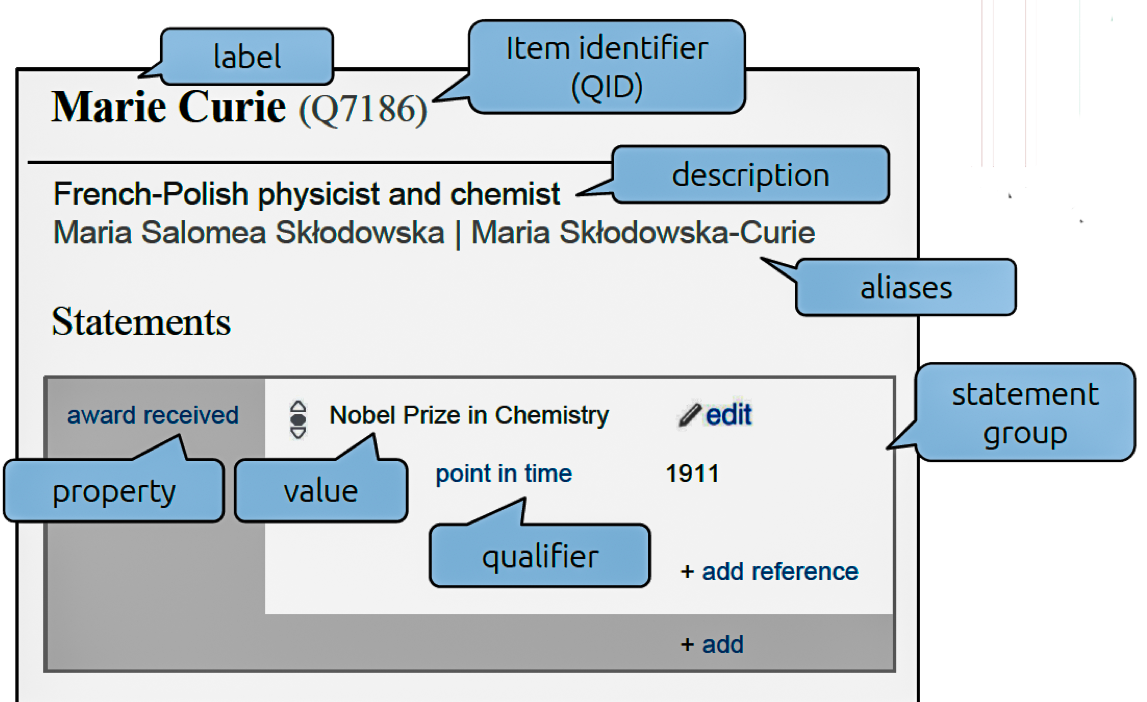}
\caption{Wikidata entities have a unique ID, a canonical name, aliases and a short description in one or more languages along with any number of statements representing properties or relations and including qualifiers and provenance metadata. }
\label{fig:wdentity}
\end{figure}

Wikidata currently has more than 900 million statements about 77 million items, supported by an ontology with nearly 2.4 million fine-grained types and more than 7,250 properties. 
An example item is the entity Q7186 shown in Figure~\ref{fig:wdentity}.  Items have a canonical name, short description and set of aliases in one or more languages. Property statements encode relations between items or between an item and a literal value and can have metadata including qualifiers (e.g., a period of time during which the property held), provenance information (e.g., the URL of an attesting source), and a rank (e.g., to distinguish a preferred value from alternative or deprecated ones). 

The data is exposed as RDF triples and can be queried using Wikimedia APIs or SPARQL queries sent to a public query service. We used the public SPARQL service to get both canonical names (e.g., Johns Hopkins University) and aliases (e.g., JHU, Johns Hopkins, Hopkins) in each of the languages studied for a given entity type (e.g., ORG). In addition, the Wikimedia community has developed many tools for searching for items, exploring the ontology, and updating entries.

\subsection{Gazetteer Generation}

The gazetteer is generated by searching Wikidata via SPARQL queries sent to the public query server to retrieve both canonical names (e.g., Johns Hopkins University) and aliases (e.g., JHU, Johns Hopkins, Hopkins)) in each of the languages studied. The first step was to construct a mapping from our project's 16 target types shown in Table \ref{tab:types} to Wikidata's fine-grained type system \cite{pellissier2016freebase}. Our types included four common core types (person, organization, geopolitical entity (GPE), location) and twelve additional types (airport, chemical, commercial organization, computer hardware/software, event, facility, government building, military equipment, money, political organization, title, vehicle).  

The mapping for some types was simple: person corresponds to Wikidata's Q5 and vehicle to Q42889.  Others had a complex mapping that eliminate Wikidata subtypes that seemed too specialized (e.g., \textit{lunar craters} and \textit{ice rumples} from Wikidata's geographic object) or allow us to retrieve more entity names given the public server's one-minute query timeout.

 \begin{figure}[t] \centering
 \includegraphics[width=\columnwidth]{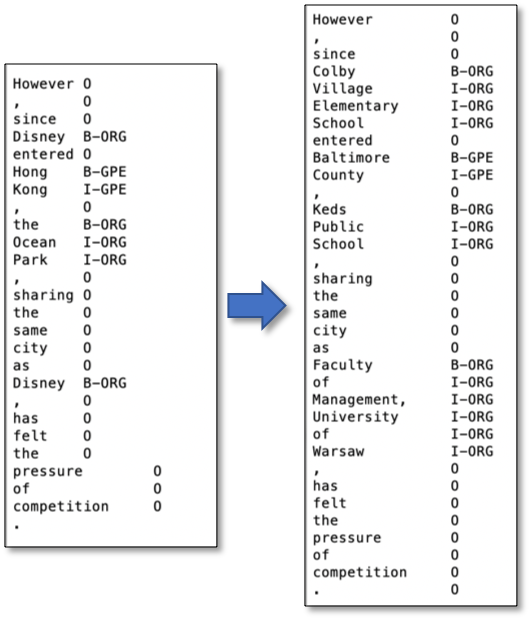}
 \label{fig:trainingdata}
 \caption{New training data is generated by replacing existing annotated names with gazetteer names of the same type.}
 \vspace{-1em}
 \end{figure}

The initial name lists were filtered by type-dependant regular expressions to delete names we thought to be unhelpful (e.g., \textit{Francis of Assisi} as a person because historical figures are unlikely to be mentioned in our targeted genres), remove Wikipedia artifacts (e.g., parentheticals), and eliminate punctuation, names that were too short or too long, and duplicate names. Although one could say that these changes bias the gazetteers, there is no reason not engineer a gazetteer in a way that is most helpful for the data. Wikidata is still being used in an automated way since we are relying on available labels. 

\begin{figure*}  
    \centering 
    \begin{tabular}{c|cccccccccccccc}
        Text & Jack & is & on & Hong & Kong & International & Airport & in & Lantau & Island & , & Hong & Kong \\
        \hline
        PER & B-PER & O& O& O& O& O&O & O& O& O& O & O & O\\
        LOC & O & O & O & O& O& O& O& O& B-LOC& I-LOC& O & O & O\\
        GPE & O& O& O& B-GPE& I-GPE& O& O& O& O& O& O& B-GPE & I-GPE\\
        ORG & O& O& O& O& O& I-ORG& I-ORG& O& O& O& O & O & O\\
    \end{tabular}
    \caption{This example shows gazetteer matches for the sentence "Jack is on Hong Kong International Airport in Lantau Island, Hong Kong" showing both full and partial matches.}
    \label{fig:gaz_ex}

\end{figure*}

We produced additional lists for Russian using a custom script that generates type-sensitive inflected and familiar forms of canonical names and aliases.  For an extreme example, the Russian name for the person \textit{Vladimir Vladimirovich Putin} ({\selectlanguage{russian}Владимир Владимирович Путин}) produces more than 100 variations. The result is a collection of 96 gazetteer files with total 15.7M  entity names, 4.2M for English, 2.1M for Russian and 584K for Chinese with an additional 8.7M Russian names produced by our morphological scripts.  We kept the gazetteers for canonical names, aliases, and inflected forms separate to facilitate experimentation.

We also worked with data downloaded directly from Wikidata, which uses a JSON serialization in which only the immediate types of each entity is provided. The entity \textit{Museum of Modern Art}, for example, is identified as an instance of an \textit{art museum}, an \textit{art institution} and a \textit{copyright holder's organisation}. To decide which, if any, of our target types an entity belongs, we constructed a dictionary that maps relevant types to our 16 target types.  For our example, this turns out to be three types: \textit{ORG}, \textit{LOC}, and \textit{FAC}.

Although doing the mapping sounds daunting given an ontology with more than two million types, it is simplified by exploiting the fact that most of these types do not have any immediate instances.  We developed SPARQL queries that identified for each of our 16 types all of their subtypes that had one or more immediate instances.  The ORG type, for example has 15,904 subtypes but only 5,962 have immediate instances;  the LOC type has only 1,181 subtypes with immediate instances.  The resulting dictionary was thus relatively small without losing any information and was used to quickly recognize entities of interest in the Wikidata dumps as well as identify their target types.

\section{Exploiting Gazetteers}

\subsection{Gazetteer Features}

To use a gazetteer as a feature in the NER system, words in the dataset are matched with a gazetteer and turned into one-hot vectors for each entity type.
Those one-hot vectors are then concatenated with word embeddings generated from other sources. For example, a word embedding of size 768 from BERT is concatenated with the gazetteer one-hot vectors sized to the {number of entities} $x$. Although each gazetteer represents an entity type, no attempt is made to communicate that type to the Bi-LSTM layer. 

We use the BIO (Beginning-Inside-Outside)
tagging scheme where \textit{B-<type>} tags the first token of an entity, \textit{I-<type>} the subsequent ones, and \textit{O} tagging non-entity tokens. For gazetteer matches, the gazetteer uses two matching schemes: full match and partial match.
\begin{itemize}
    \item Full match: an $n$-gram in the dataset matches fully with a gazetteer entry. If there are multiple matches in same entity category, the longest match is preferred.
    \item Partial match: an $n$-gram in the dataset matches partially with the gazetteer. Only partial matches of length greater than one are accepted, except for the PER type, due to frequency of one-word person names.
\end{itemize}

As an example, consider a gazetteer that contains \{Jack:PER, Lantau Island:LOC, Hong Kong Government:GPE, JFK International Airport:ORG\}, Figure~\ref{fig:gaz_ex} shows how full matches and partial matches are handled. \textit{Jack} is fully matched with PER, so it is tagged B-PER. \textit{Lantau Island} is also fully matched and tagged LOC.  \textit{Hong Kong} partially matches \textit{Hong Kong Government}, and since the match length is greater than one, it is considered a match and thus tagged as GPE. \textit{Hong Kong International Airport} is also partially matched with gazetteer entry \textit{JFK International Airport}, so \textit{International Airport} is tagged with ORG. As seen in the example match, tags of the gazetteer entries are assigned during a partial match. 
For character-level tokenized text, like Chinese, we forgo partial matching because it produces too many false matches. However for all other language, we both utilize full match and partial match as shown in the experiments.

After matching, matches are one-hot encoded with each tag type assigned a separate one-hot vector. Therefore, for each token in the text, it gets assigned a \textit{number of tag types} of one-hot vectors. These one-hot vectors are concatenated to the other features, which are fed into the BiLSTM.

\subsection{Generating Augmented Training Data}

We experimented with a second application of gazetteers that uses them to generate additional training data.
In this approach, we select sentences from our initial human-annotated training data,
replace one or more of the annotated entities with a randomly selected gazetteer entity of the same type,
and retrain the system.
However, this approach did not produce statistically significant improvements.

We developed a script that takes as input a BIO-tagged file, a type, and gazetteer for that type, and produces a modified version of the file with entity instances of the type replace with a random entity selected from the gazetteer. Additional arguments control whether all instances of a given entity in the input BIO file are replaced with the same gazetteer entity and specify the random seed, to support repeatable experiments.  Our current experiments were run by replacing entities for all types and to allowing a given input entity to be replaced with different gazetteer entities each time it appears.  Figure \ref{fig:trainingdata} shows an example with an original annotated sentence from OntoNotes on the left, and a new, generated training instance on the right.

\section{Architecture}

\begin{table} [tb]
\centering 
\begin{center}
\renewcommand\arraystretch{1.2}
\begin{tabular}{|p{10em}|p{8em}|}
\hline
\rowcolor{lightgray} \multicolumn{2}{|c|}{\textbf{\large Hyperparameters}}\\
\hline
BiLSTM layers & $1$ \\
\hline
BiLSTM hidden size & $256$ \\
\hline
BiLSTM dropout & $.5$ \\
\hline
Optimizer & adafactor \\
\hline
Gradient clipping & $1.0$ \\
\hline
Learning rate scheduler & cosine decay \\
\hline
BERT layers used & $-4,-3$, $-2,-1$ \\
\hline
Weight decay & $.005$ \\
\hline
Mini batch size & $8$ \\
\hline
\end{tabular}
\caption{Default hyperparameters used in the baseline model}
\label{tab:params}
\end{center}
\end{table}

We use a common baseline Bi-LSTM-CRF model like many sequence to sequence closed-world NER systems~ \cite{huang2015bidirectional}, 
which includes a stacked bi-directional recurrent neural network with long short-term  memory units and a conditional random field decoder and is similar to \citeauthor{chiu2016}~(\citeyear{chiu2016})
without the character-level CNN. We combine this system with BERT \cite{devlin2018bert}, which is a stack of bi-directional transformer encoders.
We keep the BERT frozen during training and testing, feeding the text into BERT and concatenating its final four layers as an input to our Bi-LSTM-CRF. 
In addition, the features generated from gazetteers are concatenated with the outputs from BERT and fed into the Bi-LSTM-CRF.
Table~\ref{tab:params} shows the hyperparameters used for our experiments. We did not perform a hyperparameter search.

\section{Experimental Data Sets}
\label{sec:datasets}

To demonstrate the effectiveness of our new approaches to NER, 
datasets labeled with names are required.
For English and Chinese, there are  
established datasets. We chose OntoNotes v5.0~\cite{pradhan2013ontonotes} 
because it has a large number of labeled entities. Table~\ref{tab:coll_stats}
contains the statistics for these datasets. 
Russian, on the other hand, has little labeled data for NER. We chose to create our own
dataset over Russian informal text. This
section describes the construction of this collection, as well
as its tag set and statistics.

\begin{figure}[t] \centering
\includegraphics[width=\columnwidth]{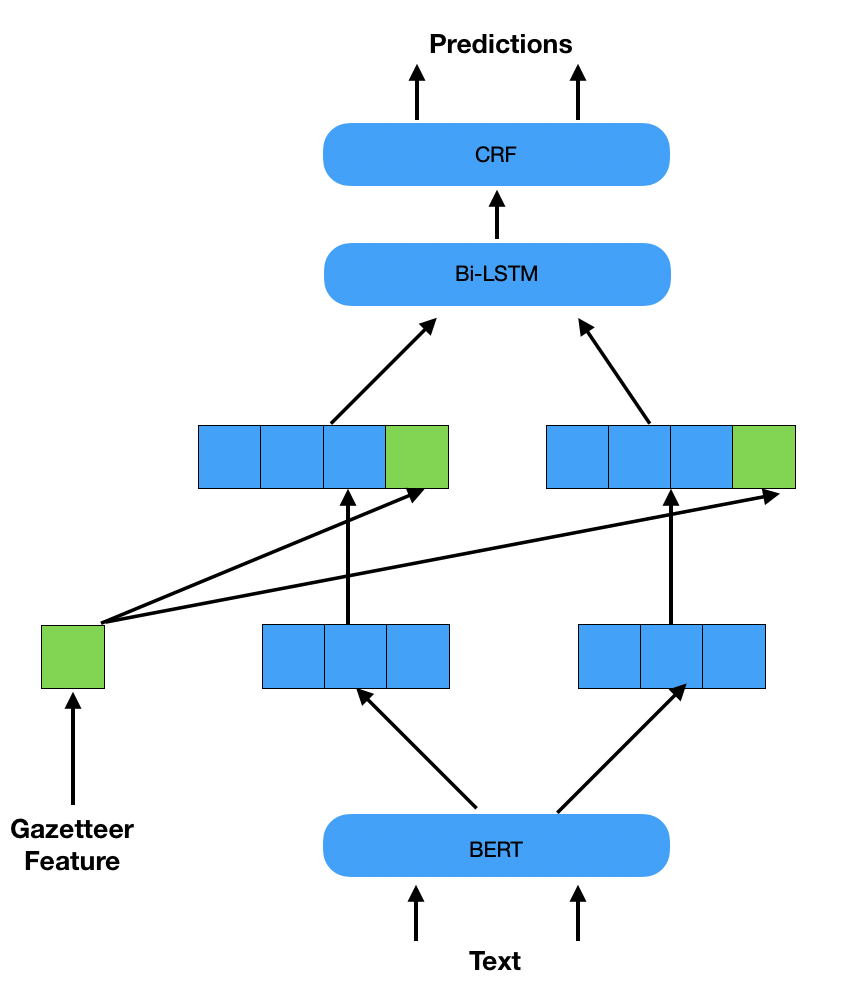}
\caption{Architecture used in our models, with baseline components in blue and additional gazetteer features in green.}
\label{fig:baseline}
\vspace{-1em}
\end{figure}

The Russian Reddit collection comes from Russian comments collected
over 433 threads on the Reddit online discussion forum \cite{reddit}.
Reddit organizes threads around a submission that is posted to a channel.
The first step in building the collection was to identify Russian threads.
Annotators examined threads with at least ten comments in a majority of Cyrillic characters to determine whether the 
thread was written in Russian.
We eliminated images and movies from the thread seeds,
as well as seeds from sites primarily devoted to image content;
such seeds typically contain few named entities in their comments.
Over 30,000 threads met these criteria. 
Threads were prioritized based on the source of the material in the 
submission, where newswire and blogs were preferred.

Annotators examined around 800 of these threads and identified the 
language of the comments, 433 of which were in Russian.  
These comments were automatically sentence-segmented using
CoreNLP~\cite{manning-EtAl:2014:P14-5}, so that named entity tagging could
be performed by annotators at the sentence level.
The \dragonfly annotation tool~\cite{dragonfly2018} was used 
to record the entity tags through an in-house Mechanical Turk-like 
interface. 

One goal of this collection was to have a wider variety of entity types so that future research could investigate types that have varying frequencies of attestation. Beyond the common core types, types were chosen that were sufficiently attested in the data. In addition, we desired to have a few subtypes of the common types to be able to experiment with this hierarchical relationship.

To assure quality annotations, either sentences
were doubly-annotated and a third annotator reviewed disagreements or 
the sentence was singly-annotated and a second annotator reviewed the
annotations. The inter-annotator agreement on the doubly annotated 
text was 53\%. The annotators agreed on whether a token was part of a name 63\% of the time. Since agreement was measured at the token level, both a name's tag and span had to match exactly. Finally the collection was split 80-10-10 into train, development, and test respectively.
Table~\ref{tab:coll_stats} shows the size of the 
collection, which was labeled with 16 types as shown in Table~\ref{tab:types}. The frequency of each type is shown in Table~\ref{tab:type_stats}. The data set is available from \url{https://github.com/hltcoe/rus-reddit-ner-dataset}.


\begin{table}[t] 
\centering \footnotesize 
\renewcommand\arraystretch{1.2}
    \begin{tabular}{|C{1.2cm}|C{1.8cm}|C{1.05cm}|C{1.05cm}|C{1.05cm}|}
        \hline
\rowcolor{lightgray} \textbf{Dataset} & \textbf{Type} & \textbf{Train}& \textbf{Test} & \textbf{Dev} \\
         \hline
          \multirow{3}{*}{\parbox{1.2cm}{English OntoNotes}} & Sentences & 82.1k & 9.0k & 12.7k \\
          & Tokens & 1644.2k & 172.1k & 251.0k \\
          & Entities & 70.3k & 6.9k & 10.9k \\
         \hline
         \multirow{3}{*}{\parbox{1.2cm}{Chinese OntoNotes}} & Sentences & 37.5k & 4.3k & 6.2k\\
          & Tokens & 1241.1k & 149.7k& 178.4k \\
          & Entities & 37.9k & 4.5k & 5.4k \\
         \hline
          \multirow{4}{*}{\parbox{1.2cm}{Russian Reddit}} &  Sentences &  22.8k & 3.2k & 3.1k \\  
          & Tokens &  281.7k & 39.3k & 37.9k\\  
          & Core Ent. & 8.1k & 1.1k & 1.0k \\ 
          & Extended Ent. & 11.2k & 1.5k & 1.4k \\ \hline
    \end{tabular}
    \caption{Statistics of dataset sizes}
    \label{tab:coll_stats}

\end{table}

\begin{table}[t] 
\centering 
\renewcommand\arraystretch{1.2}
\begin{tabular}{| l | c | c | c |}
\hline
\rowcolor{lightgray} \textbf{Tag Type} & \multicolumn{3}{c|}{ \textbf{Frequency by Collection}} \\ \hline
 & English & Chinese & Russian \\ 
 &  OntoNotes    &  OntoNotes    &  Reddit  \\ \hline 
\textbf{PER} & 27.4k    & 14.1k  & 3.3k   \\ \hline 
\textbf{ORG} & 30.0k    & 10.1k  & 1.1k  \\ \hline 
\ \ COMM     &  -       & -      & 409   \\ \hline 
\ \ POL      &  -       & -      & 174   \\ \hline 
\textbf{GPE} &  28.2k   & 20.2k  & 5.5k  \\ \hline 
\textbf{LOC} &  2.7k    & 2.7k   & 451   \\ \hline 
FAC          &  -    & -   & 50   \\ \hline 
\ \ GOVT     &  -       & -      & 36    \\ \hline 
\ \ AIR      &  -       & -      & 5     \\ \hline 
EVNT         &  -    & -   & 152 \\ \hline 
VEH          &  -       & -      & 63    \\ \hline 
COMP         &  -       & -      & 273   \\ \hline 
MIL\_G       &  -       & -      & 260   \\ \hline 
MIL\_N       &  -       & -      & 139   \\ \hline 
CHEM         &  -       & -      & 21    \\ \hline 
MISC         &  -       & -      & 1.5k  \\ \hline 
\end{tabular}
    \caption{Statistics of datasets by tag type}

    \label{tab:type_stats}
\end{table}
\section{Results using Gazetteer Features} 
\label{sec:gaz_results}

We use our models for NER tasks on the English OntoNotes, Chinese OntoNotes Russian Reddit datasets. For each, we run our baseline models and the models with added gazetteer features at least ten times, depending on the size of the collection. Smaller collections were run a greater number of times because of the greater variability in the output. Performance is reported as precision (P), recall (R), and their harmonic mean (F1).  Statistics of the dataset are shown in Tables \ref{tab:coll_stats} and \ref{tab:type_stats}. The statistics for gazetteer coverage for individual datasets are shown in Table \ref{tab:gaz_coverage_stats}. We use only the four core types for English and Chinese because our gazetteer tag types do not include the extended OntoNotes types. However, we experiment with both core and extended types for the Russian dataset. For each experiment, we train for fixed epochs and choose the model that shows the minimum loss on the development set.

\subsection{English and Chinese OntoNotes}

We use the English OntoNotes v5.0 dataset compiled for the CoNLL-2013 shared task \cite{pradhan2017ontonotes} and follow the standard train/dev/test split as presented in \cite{pradhan2013ontonotes}. We use pre-trained Cased BERT-Base with 12-layer, 768-hidden, 12-heads, 110M parameters available on the Google Github. The experiment is run for 10 trials and trained for 30 epochs. The model with the minimum dev set loss is selected and run on the test set. Table \ref{tab:en-zh-gaz} shows our experiment results. We compute the p-value of the distribution using a $t$-test. We show that adding gazetteer features increases $0.52$ F1 score, an improvement that is statistically significant ($p < 0.001$). We attribute this to an even coverage of the percentage of entities across train, dev, and test sets as seen in Table~\ref{tab:gaz_coverage_stats}, as well as a high coverage (high 80s) for GPE entities, the entity type with the largest F1 gain.

We use the Chinese OntoNotes v5.0 dataset with four core types compiled for CoNLL-2013 and follow the standard train/dev/test split as before. We use the pre-trained Chinese BERT-Base for simplified and traditional Chinese which has 12-layer, 768-hidden, 12-heads, 110M parameters. The experiment is run for 10 trials and trained for 30 epochs. The model with minimum loss on dev set is selected for testing. The gazetteer feature leads to a statically significant improvement ($p=0.003$), which we attribute this to high GPE coverage and even coverage across dataset splits. However, the absolute increase in F1 score is around 0.3, which is lower than English dataset. We believe Chinese showed less improvement due to our decision to forgo partial matches due to the high frequency of partial matched n-grams stemming from the language's logographic nature. 

\begin{table}[t]
  \centering 
  \renewcommand\arraystretch{1.2}
  \begin{tabular}{| c | c | c | c | c |}   \hline
    \rowcolor{lightgray} \textbf{Dataset} & \textbf{Model} & \textbf{P} & \textbf{R} & \textbf{F1} \\  \hline
    \multirow{3}{*}{English} &  Baseline  & 92.46 & 91.77 & 92.11 (SD: $0.10$)\\  
    & Gazetteer & 92.82 & 92.44 & \textbf{92.63} (SD: $0.12$) \\
    & +Aliases & 92.69 & 92.50 & 92.59 (SD: $0.11$) \\\hline \hline
    \multirow{3}{*}{Chinese} & Baseline  & 83.40 & 84.63 & 84.01 (SD: $0.16$)\\ 
    & Gazetteer & 83.91 & 84.72 & \textbf{84.31} (SD: $0.23$) \\
    & +Aliases & 83.84 & 84.76 & 84.30 (SD: $0.25$) \\\hline
  \end{tabular}
  \caption{Performance of BERT-BiLSTM-CRF baseline and $+$ gazetteer features on English and Chinese OntoNotes, SD stands for standard deviation}
  \label{tab:en-zh-gaz}
\end{table}

\subsection{Russian Reddit Dataset}

We use the Russian Reddit dataset to evaluate the performance of Russian NER. We use pre-trained Multilingual Cased BERT-Base with 12-layer, 768-hidden, 12-heads, 110M parameters.
We use the same baseline BERT-BiLSTM-CRF model with gazetteer feature added. For the Russian Reddit dataset, the experiment is run for 20 trials with 30 epochs. The model with minimum loss on dev is selected for testing. The different number of trials is due to the smaller size of this dataset, as is shown in Table~\ref{tab:coll_stats}. We report 
experiments with both core types and extended types using the Russian Reddit dataset. Table \ref{tab:ru-gaz} shows Russian dataset experiments with gazetteers and different tag types.

\begin{table}[t]
    \centering 
    \renewcommand\arraystretch{1.2}
    \begin{tabular}{| c | c| c | c | c | c |}
        \hline
 \rowcolor{lightgray}         \textbf{\large Dataset} & \textbf{\large Type} & \textbf{\large Train}& \textbf{\large Test} & \textbf{\large Dev} \\
         \hline
          \multirow{4}{*}{\parbox{1.5cm}{English OntoNotes}} & PER & 38.2\% & 44.3\% & 37.5\%\\
          & ORG & 19.3\% & 17.2\% & 19.0\%\\
          & GPE & 88.7\% & 86.8\% & 87.2\% \\
          & LOC & 26.3\% & 23.7\% & 30.0\% \\
         \hline
         \multirow{4}{*}{\parbox{1.5cm}{Chinese OntoNotes}} & PER & 24.0\% & 21.2\% & 21.6\%\\
          & ORG & 18.0\% & 17.4\% & 23.4\% \\
          & GPE & 76.2\% & 75.4\% & 77.2\% \\
          & LOC & 18.1\% & 17.4\% & 14.1\% \\
         \hline
          \multirow{15}{*}{\parbox{1.5cm}{Russian Reddit}} &  \textbf{PER} &  12.5\% & 15.4\% & 11.4\%                 \\  
                  & \textbf{ORG} &  16.9\% & 26.6\% &  9.8\%\\  
                & COMM &  31.4\% & 33.3\% & 5.3\%\\  
                & \textbf{GPE} &  23.4\% & 23.1\% & 20.2\%\\ 
                & \textbf{LOC} &  7.4\% & 0\% & 5.8\%\\ 
                & FAC &  4.3\% & 50.0\% & 0\%\\ 
                & GOVT &  7.7\% & 0\% & 0\%\\ 
                & AIR &  0\% & 0\% & 0\% \\ 
                & EVNT &  3.4\% & 0\% & 0\%\\ 
                & VEH &  6.1\% & 11.1\% & 20.0\%\\ 
                & COMP &  24.0\% & 17.6\% & 0\% \\ 
                & MIL\_G &  0\% & 0\% & 0\%\\ 
                & MIL\_N &  0\% & 0\% & 0\%\\
                & CHEM &  0\% & 0\% & 14.3\% \\ 
                & MISC &  0\% & 0\% & 0\% \\ 
            \hline
    \end{tabular}
    \caption{Statistics for the entity types and subtypes for each of the three collections.  Our OntoNotes data only covered the core types while our Russian Reddit included additional types and subtypes. }
    \label{tab:gaz_coverage_stats}
\end{table}

While the mean of the trials is slightly higher for those with gazetteer features, none of the results shows statistical significance. We attribute this to (a) lower coverage of our gazetteer for those in the dataset; and (b) uneven gazetteer coverage throughout train, dev, and test sets as is seen in Table~\ref{tab:gaz_coverage_stats}.  That table  reports additional results from using inflected and familiar forms of entity canonical names and aliases, as described in Section 2.  However, our takeaway here is that adding gazetteer features does not hurt the performance of the neural systems, but only improves it when the gazetteer has high coverage, as can be seen in the English and Chinese experiments.

\section{Results for Training Data Augmentation}

Using our gazetteers to produce additional training data produced mixed results and generally was inconsistent in improving performance for our models using BERT for either the four core types or the extended set.  We ran early experiments using FastText embeddings \cite{bojanowski2017enriching} and found that using our gazetteers with Russian inflections for PER improved performance for most types.  However, we did not see the same gains when using BERT.

\begin{table}[t]
    \centering
    \renewcommand\arraystretch{1.2}
    \begin{tabular}{| l | c | c | c | c |}
        \hline
\rowcolor{lightgray} \textbf{\large Tags-Model} & \textbf{\large P}& \textbf{\large R} & \textbf{\large F1} \\
%
         \hline
         C-Baseline & 80.21  & 72.12 & 75.95 (SD:0.43) \\
         C-Gazetteer & 79.81 & 72.03 & 75.72 (SD:0.44) \\
         C-Inflected & 79.75 & 72.01 & 75.68 (SD:0.42) \\
         C-Alias & 79.68 & 72.05 & 75.67 (SD:0.48) \\
         E-Baseline & 73.36 & 56.88 & 64.08 (SD:0.44)\\
         E-Gazetteer & 73.33 & 57.05 & 64.17 (SD:0.58)\\
         E-Inflected & 73.31 & 57.01 & 64.14 (SD:0.51)\\
         E-Alias & 73.08 & 57.08 & 64.10 (SD:0.48)\\
         \hline
    \end{tabular}
    \caption{Performance of BERT-BiLSTM-CRF baseline and model with gazetteer features on Russian Reddit datasets for the Core (C) and Extended (E) tag sets}
    \label{tab:ru-gaz}
\end{table}

This may be due to several reasons.  First, both core and extended types are quite broad.  Replacing the annotated ORG \textit{Harvard University} with the gazetteer ORG \textit{Disneyland} in the sentence "Professor Pinker teaches at Harvard University" seems anomalous to us and probably also to our model.  Second, our experiments were done with relatively small amounts of annotated training data, especially for Russian.  While drawing on gazetteer data may help introduce new patterns not present in the training data, such as ORGs beginning with "Association of", the chances of this helping when evaluated with the relatively small test partition is low. Third, entity names were extracted from Wikidata without regard to their utility, including both very prominent entities (e.g., the LOC \textit{Atlantic Ocean}) and very obscure ones (e.g., \textit{Avalonia}, a microcontinent in the Paleozoic era).

We plan to further explore this use case by replacing annotated entities with gazetteer entities that are in a same finer-grained Wikidata type. We can readily identify all of the Wikidata types to which a reasonably prominent entity (e.g., Harvard University) belongs.  We will select several hundred of these types as targets for gazetteer entity replacement (e.g., Q2385804 -- education institution) that are similar to an annotated entity.  We can then associate gazetteer entities with these target types, replacing an entity such as "Harvard University" with an entity that is more similar, e.g., "Swarthmore College" or "Loyola Academy").  We also plan to limit obscure entities using a measure of prominence derived from Wikidata metrics, such as their number of incoming and outgoing links.

\section{Conclusion and Future Work}

We present a simple way to generate a gazetteer, and show how it can be used in a neural NER systems.
We also present a new Russian NER corpus gathered from Reddit. 
We show that with enough coverage on the dataset, gazetteer features improve neural NER systems, even systems using deep pre-trained models such as BERT. We hypothesize that paying attention to how tuned the signal is between the gazetteer and the training set greatly impacts ho much the neural system learns to pay attention to the gazetteer. Modifying the gazetteer based on the training data is a path we plan to explored. In general, we believe gazetteer features should be a standard addition to any NER system and show that even with low coverage, the gazetteer features do not hurt the performance of neural NER systems.

While our gazetteer data augmentation did not show consistent improvement, we believe that future work in more sophisticated and contextualized replacement scheme will benefit low-resource languages such as Russian.
In addition the noisiness of the gazetteers may have a great impact on performance since the NER system may learn not to trust a gazetteer that does not assist with tagging a sufficient number of time. Future work will identify techniques to produce gazetteers that are trustworthy relative to the training data to see if such gazetteers can be shown to be more helpful.

Gazetteers and associated software is available from \url{https://github.com/hltcoe/gazetteer-collection}.

\section*{Acknowledgments} 

We thank Johns Hopkins University, the Human Language Technology Center of Excellence and the 2019 SCALE workshop for their hospitality and for facilitating an excellent research environment.

\bibliography{Bibliography.bib}
\bibliographystyle{aaai}
\end{document}